\documentclass[runningheads]{llncs}
\usepackage{graphicx}
\usepackage{subcaption}
\usepackage{amssymb,amsmath}
\usepackage{todonotes}

\newcommand{\mM}{\mathcal{M}}
\newcommand{\mP}{\mathcal{P}}
\newcommand{\mS}{\mathcal{S}}
\newcommand{\mX}{\mathbb{X}}
\newcommand{\mY}{\mathbb{Y}}
\newcommand{\ptitle}[1]{\noindent{\bf #1.}}
\newcommand{\ptitlesmallskip}[1]{\vspace{2mm}\noindent{\bf #1.}}

\begin{document}
\title{ExClus: Explainable Clustering on Low-dimensional Data Representations\thanks{The research leading to these results has received funding from the ERC under the EU's Seventh Framework Programme (FP7/2007-2013) (ERC Grant Agreement no. 615517) and under the EU’s Horizon 2020 research and innovation programme (ERC Grant Agreement no. 963924), from the Flemish Government under the ``Onderzoeksprogramma Artificiële Intelligentie (AI) Vlaanderen'' programme, and from the FWO (project no. G091017N, G0F9816N, 3G042220)}}
% \author{Xander Vankwikelberge \and Bo Kang\orcidID{0000-0002-9895-9927} \and Edith Heiter \and
% Jefrey Lijffijt\orcidID{0000-0002-2930-5057}}
\author{Xander Vankwikelberge \and Bo Kang \and Edith Heiter \and Jefrey Lijffijt}
\authorrunning{X. Vankwikelberge et al.}
\institute{Ghent University, Ghent, Belgium}
\maketitle              % typeset the header of the contribution
\begin{abstract}
Dimensionality reduction and clustering techniques are frequently used to analyze complex data sets, but their results are often not easy to interpret. We consider how to support users in interpreting apparent cluster structure on scatter plots where the axes are not directly interpretable, such as when the data is projected onto a two-dimensional space using a dimensionality-reduction method. Specifically, we propose a new method to compute an interpretable clustering automatically, where the explanation is in the original high-dimensional space and the clustering is coherent in the low-dimensional projection. It provides a tunable balance between the complexity and the amount of information provided, through the use of information theory. We study the computational complexity of this problem and introduce restrictions on the search space of solutions to arrive at an efficient, tunable, greedy optimization algorithm. This algorithm is furthermore implemented in an interactive tool called ExClus. Experiments on several data sets highlight that ExClus can provide informative and easy-to-understand patterns, and they expose where the algorithm is efficient and where there is room for improvement considering tunability and scalability.
\keywords{Dimensionality reduction \and clustering \and explainable AI \and exploratory data analysis \and hierarchical clustering \and t-SNE.}
\end{abstract}
%
%
%
%%%%%%%%%%%%%%%%%%%%%%%%%
%%%%%%%% Intro %%%%%%%%%%
%%%%%%%%%%%%%%%%%%%%%%%%%
\section{Introduction}
Artificial intelligence methods exceed human performance on many tasks and thus have found widespread use, yet the resulting models are often black boxes. There is a growing demand to create scalable human-friendly implementations. The creation of explainable white-box artificial intelligence methods is necessary to have users trust, manage, and use these in making crucial decisions \cite{xai2,xai1}.

This need is also present in the area of clustering and dimensionality reduction methods for high-dimensional data. Clustering and dimensionality reduction methods are frequently employed to get a grasp of the high-level structure of data. Especially non-linear dimensionality reduction methods such as Isomap \cite{isomap}, LLE \cite{lle}, t-SNE \cite{tsne}, and UMAP \cite{umap} manage to effectively map data onto a low-dimensional space, retaining the relative distances from the high-dimensional space, but this low-dimensional space is often difficult to understand \cite{sedlmair2012}.

\ptitle{Related work}
Possible solutions in several directions have been studied: we may explore the 2D projection by letting the data glyphs correspond to specific attributes (e.g., color the points using the values for a specific attribute) or draw attribute isolines (as in DimReader \cite{faust2018}). Such solutions are limited to a single or very few attributes at a time and it is not obvious how to make them practical for data whose original dimensionality is large (e.g., from 20 attributes or more).

There also exist approaches that focus on data points rather than attributes: `Forward projections' show how points would move in the low-dimensional embedding if their attributes would change, and we may also do the inverse exercise of `backward projection', i.e., how the attributes would need to change to move a point in a certain direction (which does not have a unique solution, so an inductive bias is necessary to resolve this) \cite{cavallo2018chi}. However, it is not practical to learn the structure of a large data set by exploring each point individually. `Probing clusters' \cite{stahnke2016} means to explore the feature values for manually selected sets of points that for example appear to form a cluster. This may lead to insights on the high-level structure. Yet, for high-dimensional data we are still left with the problem that this does not scale to a large number of attributes.

Most similar to our work is the Clustrophile 2 tool \cite{cavallo2018vis}, which lets users explore a diverse set of pre-generated clusterings by means of a scatter plot showing a low-dimensional projection, as well as an attribute similarity matrix and/or a decision tree for the cluster assignment. However, the decision tree does not explicitly show how each cluster stands out, while the similarity matrix is only a visual aid that puts the burden of the comparison fully on the user, who are likely quickly overwhelmed for data with a large number of attributes.

\ptitle{Contributions}
We propose in this paper to take a very different approach: to compute an interpretable clustering that facilitates the analysis of a scatter plot of dimensionality-reduced data. We do this in several steps: (1) first we use information theory to quantify the informativeness of the mean and variance statistics for a given subset of points and given subset of attributes. (2) To control the trade-off between the complexity  of the clustering and the amount of information gained, we introduce a simple notion of complexity for such a `bicluster pattern', which gives us control of how complicated the explanations for the clusters may be. (3) Integrating this with an approach to cluster the data on the low-dimensional representation, we obtain coherent clusters on the dimensionality-reduced scatter plot along with a subset of attributes that are informative (in the information-theoric sense), to explain this cluster. (4) Using this, we attempt to automatically generate an optimal explainable clustering on low-dimensional representations of data sets, so that users can meaningfully explore complex data with a limited time cost for interpreting apparent structure.

\begin{figure}[h]
\centering
\includegraphics[width=0.94\textwidth]{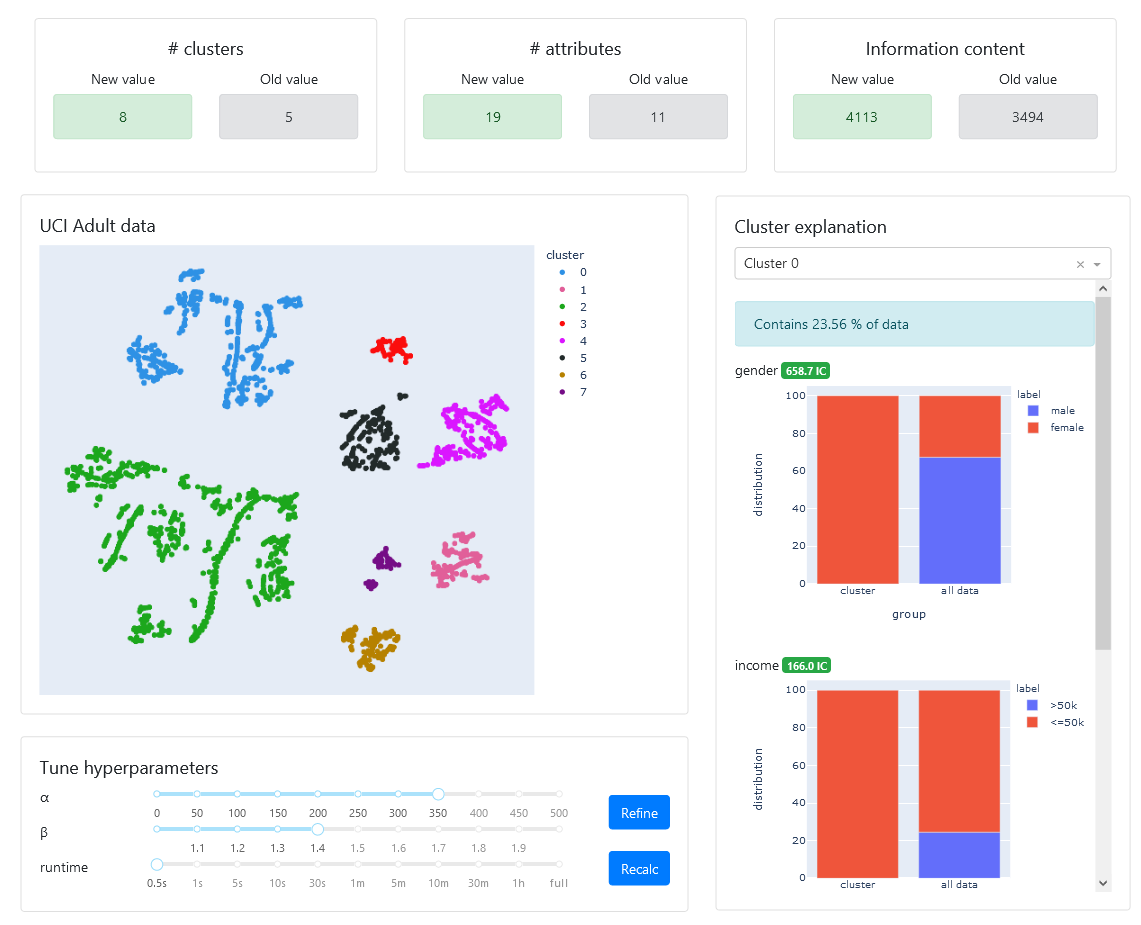}
\caption{The ExClus user application. Results are generated by applying the algorithm on the UCI Adult data set, an extract from the 1994 USA census \cite{uci}.}
\label{fig:dash_app}
\end{figure}

The contributions of this paper are as follows:
\begin{itemize}
    \item We define bicluster patterns, their informativeness using information theory and a simple notion of their interpretational complexity.
    \item We derive an algorithm to optimally cluster a dimensionality-reduced scatter plot such that the clusters are as informative as possible using a few attributes in the high-dimensional space.
    \item We study the computational complexity of this problem and argue for the use of hierarchical clustering to reduce the search space of possible solutions.
    \item We provide an implementation of that algorithm as well as a browser-based tool called ExClus (see Fig.~\ref{fig:dash_app}) that can be used to explore data in practice.
    \item We present experimental results on a few datasets to explore the usefulness of the tool as well as the effect of the two hyperparameters that govern the interpretional complexity of bicluster patterns.
    We present some early feedback from users and empirically study the scalability of the method.
    \item We find that ExClus has great potential to further facilitate data exploration with dimensionality-reduction methods and that the method is sufficiently scalable to use also on large data.
\end{itemize}
The ExClus tool is freely available at https://github.com/aida-ugent/ExClus 

The paper is structured as follows: in Section~\ref{sec:method} we first formalize the type of patterns that we aim to extract and define the informativeness and descriptional complexity of these patterns. We then introduce the optimization algorithm to extract the patterns. In Section~\ref{sec:exclus} we introduce ExClus and the user interface. Experiments are given in Section~\ref{sec:experiments} and conclusions are presented in Section~\ref{sec:conclusions}.

% \todo[author=Jef,inline]{I write above that we introduce `bicluster patterns'; this is just a suggestion, I think it would be useful to give the object that is returned by the algorithm a name, but please let me know what you think about this Xander. If you are fine with this I can adjust the writing accordingly after you have made your pass this weekend.}

% \todo[author=Xander, inline]{Bicluster patterns sounds appropriate, and I do like introducing it as it will make the text easier to write and read.}

%%%%%%%%%%%%%%%%%%
%%%% Method %%%%%%
%%%%%%%%%%%%%%%%%%
\section{Method\label{sec:method}}
% An automated tool for generating explainable clusters relies on creating an efficient algorithm that succeeds in this regard.

% \subsection{Idea}

The general idea is to define the substructure that we aim to find as a pattern, which we call a \emph{bicluster pattern}. Informally, this is a subset of points and the mean and variance statistics for a subset of the attributes. The approach then builds upon the framework for subjective interestingness for patterns \cite{debie2013ida}, which suggests the use of information theory to quantify how much information a user gets per time unit spent. There are three key concepts in this framework: information content, description complexity, and subjective interestingness.

The \emph{information content} expresses how much the user learns by showing them a specific pattern. The \emph{description complexity} aims to express the difficulty of understanding the pattern, and the \emph{subjective interestingness} is simply the ratio of the two. The term `subjective' explicates that how much we learn can only be specified with respect to prior expectations over the data. In practice we will use a prior based on the data, but it may also be chosen subjectively. Barring issues of feasibility, the prior could reflect the actual knowledge of a user.

We discuss each of these concepts in more detail: In Section~\ref{sec:bicluster}, we express the \emph{information content} of a bicluster pattern, i.e., the number of bits of information that we learn by showing the statistics for this bicluster, in comparison to given prior expectations. In Section~\ref{sec:descriptioncomplexity}, we also introduce the \emph{description complexity} that aims to capture how difficult or time consuming it is to process the presented information, for a human end-user. In Section~\ref{sec:exclusproblem}, we then formalize the \emph{explainable clustering problem}: to find a set of bicluster patterns that partition the data into clusters that are coherent with respect to a given 2D projection, such that the \emph{subjective interestingness} of the clustering is maximized.

\subsection{Bicluster patterns and their information content\label{sec:bicluster}}

\ptitle{Notation} Let $\mX$ be an $n\times m$ data matrix with $\mX_i$ denoting data point $i \in \{1, \ldots, n\}$ and $\mX_{ij}$ the value for the $j$-{th} attribute ($j \in \{1, \ldots, m\}$) of $\mX_i$. We write $t(j) \in \{\textrm{bool},\textrm{real}\}$ to denote the type of attribute $j$.

\begin{definition}
A bicluster pattern $\mP$ is a tuple $\left(D,A,\mS\right)$ with $D \subseteq \{1, \ldots, n\}$ a set of data points indices, $A \subseteq \{1, \ldots, m\}$ a set of attribute indices, and the statistics $\mS = \{\mS_{A_1},\ldots,\mS_{A_{|A|}}\}$, corresponding to the attributes whose indices are in $A$. For boolean attributes ($t(j) = \textrm{bool}$), $\mS_j \in [0,1]$ is a frequency and for real-valued attributes ($t(j) = \textrm{real}$) both a mean and standard deviation: $\mS_j \in \mathbb{R} \times \mathbb{R}^+$.
\end{definition}

To express how much we learn by observing the statistics for a set of attributes for a subset of the data, we first need to express what we are comparing against, i.e., we have to define a prior distribution for the data. We take a simple approach here and use as prior the maximum likelihood statistics fitted on the full data, without accounting for any co-variate structure. That is, each attribute is assumed to be independent. Although in some context it may be preferable to already account for co-variates between the attributes, in many cases the independence assumption is good as it is also transparent for the user.

Similarly, the information that we obtain by observing a bicluster pattern are the maximum likelihood statistics $\mS$ for the set of points $D$ and the attributes $A$. The statistics may be used to derive a Maximum Entropy model for the data. As the statistics are assumed independent, indeed also the Maximum Entropy model is independent and we may write it for each attribute separately. It depends on the attribute type, so for brevity we write this model as $\mM_{t(j)}(\mS_j)$. The Maximum Entropy model is a Gaussian distribution for real-valued attributes and a Bernoulli distribution for boolean attributes.

The information content of a pattern $\mP$ is equivalent to the Kullback-Leibler divergence between the prior expectations and the statistics contained in $\mP$:
\begin{align}
    I(\mP) = \sum_{i=1}^{|D|} \sum_{j=1}^{|A|} D_{KL}\left( \mM_{t(A_j)}(\mS_{A_j}^D) || \mM_{t(A_j)}(\mS_{A_j}^{\mX}) \right).
\end{align}
Note that we overloaded $D_{KL}$ here to refer to the KL divergence, but in all other occurrences $D$ indeed refers to a subset of the data points included in a bicluster pattern. The KL divergences are straightforward to compute analytically for both the Bernoulli and Gaussian distribution.

Note that it may happen that the variance of a bicluster for a specific real-valued attribute is zero, in which case the KL divergence will be infinite. Therefore, we add a small value $\epsilon$ to all variance estimates. Resolving this in a more robust manner by for example considering the precision of the real-valued numbers is left for future work.

\subsection{Description complexity\label{sec:descriptioncomplexity}}

The aim of the description complexity is to quantify how difficult it is to process the presented information, i.e., how time consuming it is to internalize. Unfortunately we have no realistic models of human cognition so we will just need to work on assumptions. It is important to realize that our aim here is not to do model selection in the statistical sense and we are not presenting the patterns to another computer. The aim is simply to have a formula that is suitably parameterized such that we can balance the amount of information and the complexity of the identified patterns.

In previous papers on subjective interestingness, the description complexity was quantified as a linear function over the number of statistics that is presented to the user. Often, this leads to a more tractable optimization problem. We instead choose to make it explicit that providing more statistics realistically has a superlinear effect on the amount of time to process the information. Because we are going to present the user not a single bicluster pattern, but a set that partitions the data, we define the description complexity directly for a set of bicluster patterns as $\alpha$ plus the total number of statistics to the power $\beta$:
\begin{align}
    C(\{\mP_1,\ldots,\mP_k\}) = \alpha + \left( \sum_{i=1}^{k} \sum_{j=1}^{|A^{\mP_i}|} |\mS_{A_j^{\mP_i}}| \right)^\beta.
    % \sum_{i \in l}^{} \sum_{j \in S_{i}}^{}c_{I_{i},j}}{\alpha+(\sum_{i \in l}^{} \sum_{j \in S_{i}}^{}d_{j})^{\beta}}.
\end{align}
Here $|\mS_{A_j}| = 1$ for boolean attributes and $|\mS_{A_j}| = 2$ for real-valued attributes. This formula also includes two hyperparameters, $\alpha$, and $\beta$, which allow for tuning by users and enable the identification of more intuitive solutions.

% A formal version of the descriptional complexity gets its roots from the minimum description length, which formalizes Occam's razor. Occam's razor assumes the simplest explanation (shortest description length) is the best one, and this perfectly aligns with the above-mentioned intuitive explanation for descriptional complexity. The description length also differentiates between attributes as binary attributes are more straightforward to represent than real-valued attributes. For this reason, binary attributes receive a description length of one and real-valued attributes of two.

\subsection{Explainable clustering problem\label{sec:exclusproblem}}

Finally, we want to find a clustering that makes a trade-off between the total information content and description complexity. Hence, we define the subjective interestingness for a set of bicluster patterns as 

% to give each clustering, combined with its explanation, a value that it can then use to compare different possibilities and select the best one. This trade-off, which enables the optimization and is named the subjective interestingness (SI) of a clustering, is conveniently represented in Formula \ref{SI_general} as the ratio of information content (IC) and descriptional complexity.

\begin{equation}
    S(\{\mP_1,\ldots,\mP_k\}) = \frac{\sum_{i=1}^k I(\mP_i)}{C(\{\mP_1,\ldots,\mP_k\})}.
    \label{SI_general}
\end{equation}

The missing component is that we have not related these concepts yet to the low-dimensional projection that we started out from. Let $\mY$ denote this low-dimensional---typically 2D---projection of $\mX$. We may now define the explainable clustering problem.
\begin{problem}
The explainable clustering problem is to find a set of bicluster patterns $\{\mP_1,\ldots,\mP_k\}$ ($k \geq 1$) that partition the data (i.e., they cover all data points $\bigcup_{i=1}^k D_i = \{1,\ldots,n\}$ and no data point is covered twice $D_i \cap D_j = \emptyset \ \forall i \neq j$) in a coherent manner with respect to $\mY$ and that maximize the subjective interestingness $S(\{\mP_1,\ldots,\mP_k\})$.
\end{problem}

Here we have not defined exactly when a partitioning may be called coherent with respect to $\mY$. Indeed, this may differ per usage scenario and we argue that it is not obvious there may exist a universally applicable definition. We will argue for a particular choice in the following section that also benefits the design of an efficient optimization algorithm.

\section{Search algorithm}

We are faced with a difficult optimization problem: we want to cluster the data in a given low-dimensional projection, in such a manner that the attribute values are coherent and we are concurrently selecting attributes to explain the clusters.

We first considered a two-step approach; first clustering the data and then computing the optimal explanations. Alternatively, we could derive an iterative optimization scheme, e.g., an EM scheme switching between data point assignment and the explanation attributes. However, we opt for a third solution where we constrain the search space to make it feasible to optimize the cluster assignment, number of clusters, and the attribute selection in an integrated manner.

% It is unlikely that it would be possible to solve this formalized problem optimally efficiently due to the enormous search space for the maximization problem. To circumvent this scalability issue, the implementation of the algorithm uses heuristics to speed things up. These approaches result in the algorithm not reaching a proven global optimum for Formula \ref{optimization_hyper}. However, it can get to several local optima very fast.

% \ptitlesmallskip{Clustering method}
% The number of possible clusterings is already enormous and increases with the number of data points. Luckily, many methods exist for clustering data which can help reduce the search space. So, instead of going through all possible patterns in a brute force way, the algorithm can compare different outputs from an existing method.

\subsection{The ExClus algorithm}
\label{algorithm}
\ptitle{Step 1} We start by computing a hierarchical agglomerative clustering using the Euclidean distance on the projection $\mY$. A tested and popular method for clustering data, this conveniently ensures coherence of the final clustering solution and it constrains the search space. The resulting dendrogram will be used to guide the full clustering, but we will not use the distances in $\mY$ to select clusters.

\ptitle{Step 2} Given the optimization problem, we have to consider how to cut the dendrogram to obtain the best set of bicluster patterns. We are still faced with a large number of possible clusterings. If the tree is sufficiently large and balanced, the number of possible clusterings with $k$ clusters is given the corresponding Catalan number, which is already 1430 for $k=8$, for example.

% \ptitlesmallskip{Greedy search strategy}
% Although the introduction of hierarchical clustering dramatically reduces the search space, there are still many possible options to compare when only considering the hierarchical clustering tree.  Furthermore, the tree also increases rapidly with data set size, as each extra data point adds another leaf node to the dendrogram.

To obtain an approximate solution, we optimize the clustering by splitting one cluster of at a time. That is, we start with everything in one cluster, then consider all possible splits for $k=2$. This can be done in time linear in the size of the dendrogram, as indeed we may exactly split any branch off. For each possible split, we can optimize the attributes by ranking them and greedily considering whether we should add a boolean or a real-valued attribute (because they weigh differently in the description complexity). We obtain indeed the optimal solution for $k=2$ given the constraints, but then we take the $k=2$ solution and split this once more in any position in the tree to obtain a solution for $k=3$, which is not guaranteed to be optimal. The procedure continues for $k > 3$.

% The greedy search idea is to select one clustering, with the highest subjective interestingness (Formula \ref{optimization_hyper}), for each number of clusters and compare each of them to select the optimal number of clusters eventually. The approach is greedy because it starts with one cluster, then two, and so on, and each time the optimal selection for a specific number of clusters starts from the optimal clustering with one cluster less.

This procedure could go on until the number of clusters is equal to the number of samples. This would be time consuming and it is highly unlikely that a solution with very large $k$ would maximize the subjective interestingness. Hence, we implement a time limit to ensure the runtime stays in an acceptable range. This limit can be any value, from a few milliseconds to only stopping when all calculations end.

\subsection{Solution refinement\label{solution_refinement}}

It is not obvious how to tune the hyperparameters, so we expect this to be done through trial-and-error. From experiments presented in Section~\ref{sec:experiments}, we find they can have a dramatic effect on the resulting solution and the differences may also be abrupt. Restarting the algorithm from scratch is then not always desirable to find a good solution. Hence, we also introduce a procedure to modify the current solution given new hyperparameter values.

% The introduction of hyperparameters in combination with an efficient search algorithm enables users to rapidly compare different solutions. However, restarting the algorithm from scratch for each parameter change can result in confusing outcomes.

% Every time the algorithm restarts it again starts at one cluster and builds the next possible clustering by splitting one of the clusters from the optimal clustering with one cluster less. So, for example, if the optimal number of clusters is eight before and after the hyperparameter change, the result can still be completely different because each step of the greedy search can select a different clustering as optimum.

% Therefore, the algorithm also includes an option to start calculations for hyperparameter changes from the current clustering and evaluate if merging or splitting some clusters, according to the dendrogram, results in a new optimal solution. This will result in more intuitive outcomes.

This modification again uses the dendrogram, but instead of starting at one cluster, it starts from the previously obtained result. The procedure then evaluates both whether splitting more clusters off, as described in Section \ref{algorithm}, or merging some clusters back together that were previously split off results in a higher subjective interestingness.

%%%%% Application %%%%%%
\section{User application\label{sec:exclus}}
In this section we present the user interface of ExClus, which is based on the previously introduced formalization and algorithm. The application allows testing and intuitive usage of the presented algorithm. In the following, we highlight different aspects of the interface and summarize user feedback. 
%Evolving from a mathematical formalization and optimized implementation, ExClus is provided as an application so it may be tested and used more extensively.

\subsection{Design choices}
The interactive tool helps users to understand data sets by generating clusters and their explanations simultaneously. Users can browse through through the explanations at their own pace and tune the algorithm to retrieve more insights. The application is created with Dash, a Python framework for building data science web applications. Figure~\ref{fig:dash_app} shows the user interface.

\ptitlesmallskip{Dashboard}
The dashboard, situated at the top of the application, shows some essential information on the currently presented clustering. After tuning the hyperparameters to get different results, it also shows how these values changed. Alone, these values do not tell much, but after tuning, it is helpful the compare how they changed.

\begin{itemize}
    \item \textbf{\# clusters:} How many clusters the algorithm generated.
    \item \textbf{\# attributes:} Each cluster needs a certain number of attributes as an explanation (at least one). For example, cluster one might need only one attribute, but cluster two might need four. This number is the sum of all these attributes for all clusters.
    \item \textbf{Information content:} This is the entire clustering's Information Content\footnote{It is a deliberate choice to show the Information Content and not the Subjective Interestingness score, as the latter depends on the values of the hyperparameters. It is not sensible to compare those values across different hyperparameter choices, while the Information Content can indeed be compared.}.
\end{itemize}

\ptitlesmallskip{Data Embedding}
%This graph shows the clustering result from the ExClus algorithm on the low-dimensional data representation. 
In the top left of the interface we show a scatter plot of the low-dimensional data and display the clustering result from ExClus with different colors. Going from multiple dimensions to only two with a non-linear transformation, which dimensionality reduction techniques as t-SNE \cite{tsne} do, results in difficult to interpret axes, so the only thing deductible from the graph, besides the clustering, is that points close together on the 2D plot are similar in the high-dimensional space.

\begin{figure}[t]
  \centering
  \begin{subfigure}{0.5\linewidth}
    \centering
    \includegraphics[width=0.90\linewidth]{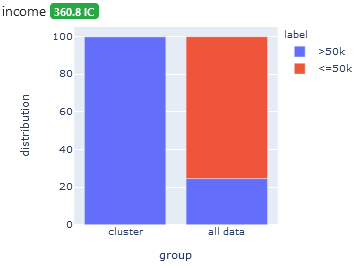}
    \caption{Explanation of a binary attribute.}
    \label{explain_cluster_4:binary}
  \end{subfigure}%
  \begin{subfigure}{0.5\linewidth}
    \centering
    \includegraphics[width=0.90\linewidth]{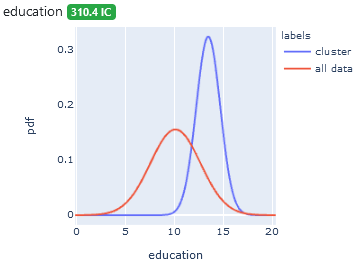}
    \caption{Explanation of a real-valued attribute}
    \label{explain_cluster4:non-binary}
  \end{subfigure}
  \caption{Part of the explanation for cluster four in Figure \ref{fig:dash_app}.}
  \label{fig:explain_cluster4}
\end{figure}

\ptitlesmallskip{Cluster explanations}
This part of the application provides the identified explanations of why specific points form a cluster. When users select one cluster using the drop-down menu, we show the relative size of the cluster and the description of attributes that make this cluster distinct from the rest of the data. We sort the attributes by decreasing information content and use different visualizations for binary and real-valued attributes as shown in Figure \ref{fig:explain_cluster4}.
%This part of the application provides the much-needed explanation of why specific points form a cluster. Through the drop-down menu, users can select which cluster to see explained. A cluster description exists of two parts: a box gives information on the percentage of data included in the cluster, and it presents information on the attributes needed to explain the cluster. The two types of attributes (binary, real valued) are explained in a different manner.

For binary attributes, we illustrate the distribution of values in a stacked bar chart. The left bar shows the distribution within the selected cluster and the right bar the distribution of the entire data set. For real-valued attributes, we derived the information content by fitting Gaussian distributions using mean and standard deviation as parameters. To compare the selected cluster to the entire dataset, we plot both probability density functions for the Gaussian distributions in the prior model and the model of the cluster.
%Next to the stacked bar for the selected cluster and attribute is a stacked bar chart that gives the distribution of said attribute for the entire data set (i.e., this shows the prior expectation).
%Just as during calculations of the KL divergence the real-valued attributes are assumed to follow a Gaussian distribution where the mean and standard deviation of the attribute are used as parameters. In order to compare the selected cluster to the entire data set, both probability density functions are plotted.
%As a final part of the explanation, the application shows each attribute's information content. These values are displayed next to the attribute title with a green badge because they inform users of how important an attribute is.

\ptitlesmallskip{Hyperparameter tuning}
Below the data visualization we display the current parameter values that influence the number of clusters and the detailedness of their explanations. Range sliders allow the user to change $\alpha$, $\beta$, and the runtime limit of the greedy search. A more extensive look into the effects of these parameters follows in Section \ref{sec:experiments}.
%The last part of this application, the hyperparameter tuning, allows users to influence the number of clusters and the depths of the explanations. A user can change three parameters: the $\alpha$ and $\beta$, and also the runtime of the greedy search algorithm.
%
%Increasing $\alpha$ results in more detailed explanations and more clusters. On the other hand, an increase in $\beta$ has the opposite effect as $\alpha$. Furthermore, $\alpha$ has a more gradual effect, while $\beta$ can have a more abrupt effect. A more extensive look into the effects of these parameters follows in Section \ref{sec:experiments}. The runtime is also included as a parameter because the algorithm can run for a long time while it tries to check the entire hierarchical clustering tree.
%
Next to the parameter sliders we provide two ways of applying the new parameter values: \textit{refine} and \textit{recalc}. The refine option starts the algorithm from the current clustering as described in Section \ref{solution_refinement}. The recalc option starts the algorithm from scratch. That way, users can either build on the current clustering and understanding of the data or investigate a new approach.

\section{Experiments\label{sec:experiments}}

In this section we discuss an empirical evaluation that we conducted to test ExClus. First we present the results from case studies on three data sets, secondly we consider quantitatvely the effects of the hyperparameters, and finally we discuss experiments on the scalability of the method.

%%%%% Use Cases %%%%%%%
\subsection{Use cases}
This section gives a short evaluation of the algorithm's results to give an initial insight into the application's possibilities.

\ptitlesmallskip{UCI Adult} The UCI Adult data set is an extraction from a 1994 USA census database, and the attributes included are: age (continuous), gender (male/female), ethnicity (white/other), education level (continuous), hours worked per week (continuous), and income (${\leq}50$k, ${>}50$k). This experiment uses a sample of only 2500 of the nearly 32000 data points for performance reasons. An example of the algorithm's result is visible in Figure~\ref{fig:dash_app}.

Dissecting the clustering reveals that the dimensionality reduction method mostly used three attributes to create the clusters: gender, income, and ethnicity. However, by tuning the hyperparameters, the algorithm can split these clusters further to reveal more details on the people included in each group or put clusters back together to show they are part of a more general pattern. For example, cluster two (green) contains all the white males with an income above \$50k, which is more than 30\% of the data set. There is a possibility of splitting this cluster further to distinguish between attributes such as education level or hours worked per week by reducing $\beta$. However, this can overwhelm a user initially, so it is better to start with few clusters and then refine them to learn more details.

Other significant sections in this clustering are cluster zero (blue), which are the female counterparts of cluster two, and this cluster's explanation is also partially shown in Figure~\ref{fig:dash_app}. The top right then contains all the high-income white people, with the red cluster being females and the other two (black and magenta) males, where the algorithm further separates them on high education level (black) and average education level (magenta). This difference also reveals that the dimensionality reduction does not reveal every pattern because there is no such split for the females as there are not enough high-income females included for the dimensionality reduction method to emphasize it. Finally, the bottom right has three clusters, including only other than white ethnicities, and the algorithm split them in almost the same manner as the white people, but again with less detail as there are not that many of them in the data sample.

\ptitlesmallskip{German socio-economic data}
This data set includes the socio-economic information of 412 German districts \cite{boley2013}. There are more than thirty attributes, but many of them are related, and overall, there are three main categories. First, the voting record attributes contain the voting percentages for the five largest parties (Green, Left, CDU, SPD, and FDP) in the 2005 and 2009 elections, where we included only the latter attribtes. Another block of attributes define the age distribution for the district, such as the percentage of old or young people. The remaining attributes give information on the workforce, such as which percentage of people work in which sector (agriculture, finance, service, etc.).

\begin{figure}[t]
  \centering
  \begin{subfigure}{0.5\linewidth}
    \centering
    \includegraphics[width=0.9\linewidth]{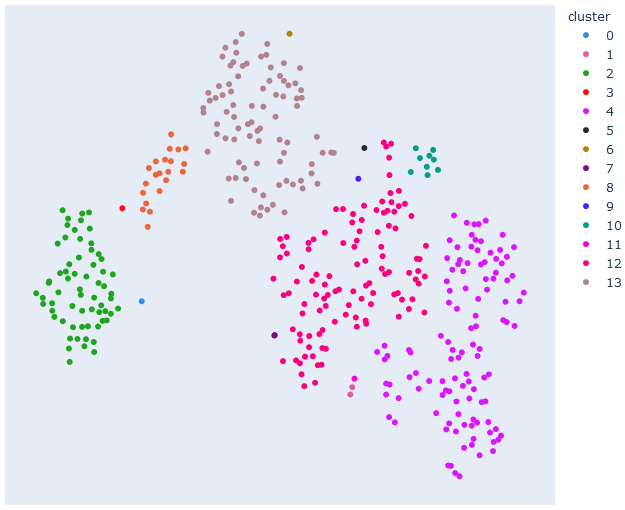}
    \caption{Single linkage}
    \label{german_clustering_sl}
  \end{subfigure}%
  \begin{subfigure}{0.5\linewidth}
    \centering
    \includegraphics[width=0.9\linewidth]{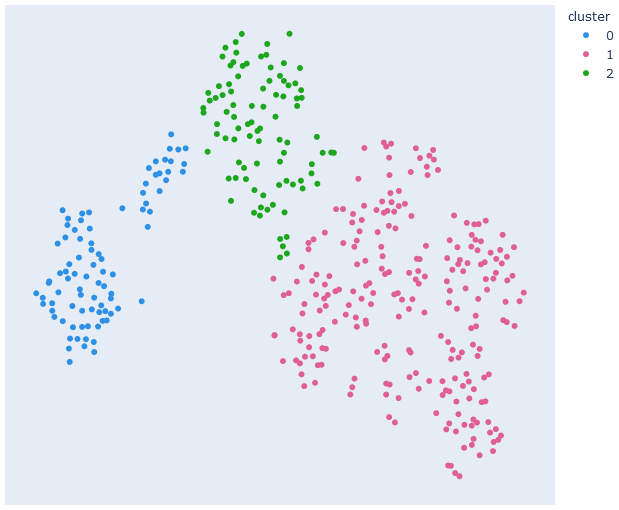}
    \caption{Complete linkage}
    \label{german_clustering_cl}
  \end{subfigure}
  \caption{Results from the ExClus algorithm on the German socio-economics data for hyperparameter values $\alpha$ 250, $\beta$ 1.6 and different linkage criteria.\label{german_clustering}}
\end{figure}

The application's outcome for two different parameter settings, shown in Figure~\ref{german_clustering}, highlight a challenge in making ExClus effective. Notably, for hyperparameter settings that gave solutions with only a few clusters on other data, the solution here has two clusters with only a tiny number of data points (cluster zero and one). On the UCI Adult data this happened only with extreme parameter settings. Changing the hyperparameters here worsens this issue. On the one hand, a user could solve this by changing the hyperparameters of the dimensionality reduction method, t-SNE in this case, but it can be a strenuous task to test different embeddings on the application each time. Furthermore, assuming that the low-dimensional representation is given, we would prefer that the application deals with this problem regardless of the embedding. It appears the problem is at least partly related to the dendrogram of the agglomerative hierarchical cluster. Specifically, the problem disappears if we consider another linkage criterion, where we used single linkage in other cases, here complete or average appears to be preferable. The single linkage appears to fail here because the clusters are not as homogeneous as for other data sets.
% The clusterings generated with the complete linkage criteria, as seen in Figure \ref{german_clustering_cl}, show more promising results as it has the benefit of not creating an overload of tiny clusters.

\begin{figure}[t]
    \centering
    \includegraphics[width=0.9\textwidth]{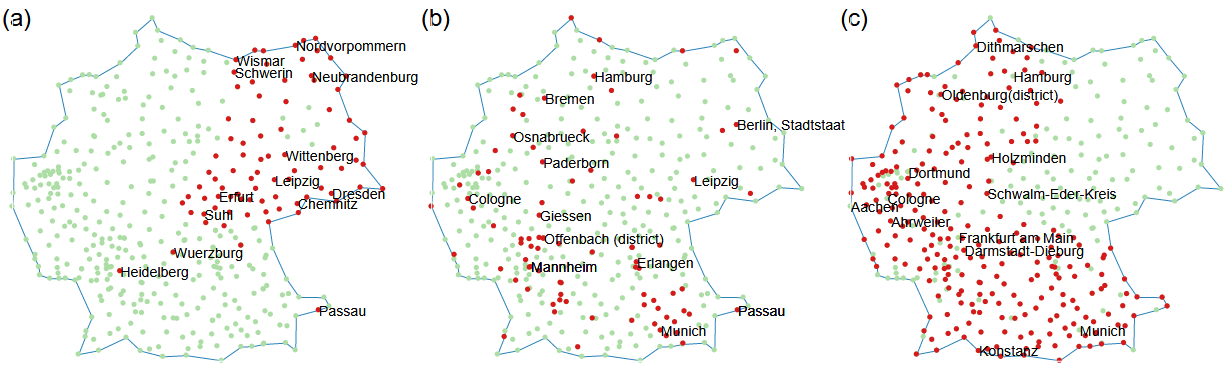}
	\caption{Patterns discovered on the GSE data in the subjectively interesting subgroup discovery paper (Lijffijt et al. \cite{lijffijt2018subjectively}). % (a) Low numbers of children are present in Eastern Germany, as well as in three cities with a very high percentage of students (Heidelberg, Passau, Wuerzburg). Here the Left party is popular is popular at the expense of all other parties. (“Children Pop. \textless= 14.1”) (b) These are larger cities with relatively many jobs. Here the Green party is more popular at the expense of Left.("Middle-aged Pop. \textgreater= 26.9"), (c) These are cities with many children. Here the Green party is more popular at the expense of Left. ("Children Pop. \textgreater= 16.4"). 
	Reproduced with permission.}
	\label{fig:germany_maps}
\end{figure}

Interestingly, looking at the clusters in Figure~\ref{german_clustering_cl}, and highlighting the districts on a German map, they align almost entirely with the patterns discovered in previous research by Lijffijt et al. in their research on finding subjectively interesting subgroups in data sets with real-valued attributes \cite{lijffijt2018subjectively}. Comparing some of the attributes included in the explanations of the clusters further proves that these patterns are almost equal. Take, for example, cluster two (green), which maps to pattern b in Figure \ref{fig:germany_maps}. Just as in the paper, this cluster consists of districts with a higher population density, indicating they are cities, with a political favor towards the Green party and a large percentage of middle-aged people. 

\ptitlesmallskip{Cytometry data}
Cytometry measures the characteristics of cells and has a broad range of applications. In this case, the experiment uses single-cell data of one mouse also used in previous research, which looked at different ways to make sense of increasingly higher dimensional data retrieved with a specific cytometry technique \cite{cytometry}. Each cell is described by nine markers and the researchers in question used dimensionality reduction and clustering methods, which makes it ideal to evaluate ExClus and compare results.

\begin{figure}[tp]
    \centering
    \begin{subfigure}{0.5\linewidth}
        \centering
        \includegraphics[width=0.77\textwidth]{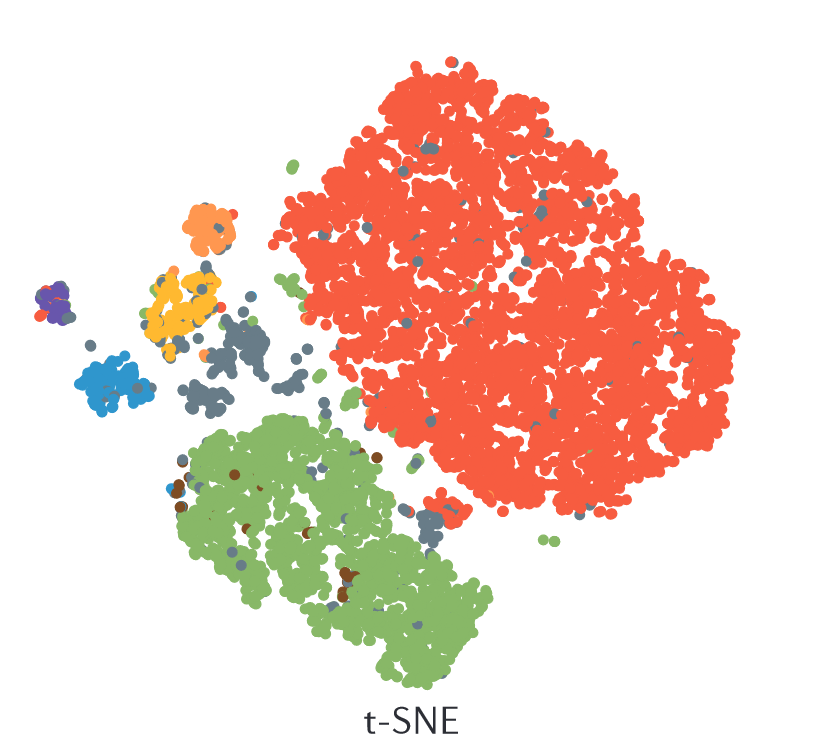}
    \end{subfigure}%
    \begin{subfigure}{0.5\linewidth}
        \centering
        \includegraphics[width=0.77\textwidth]{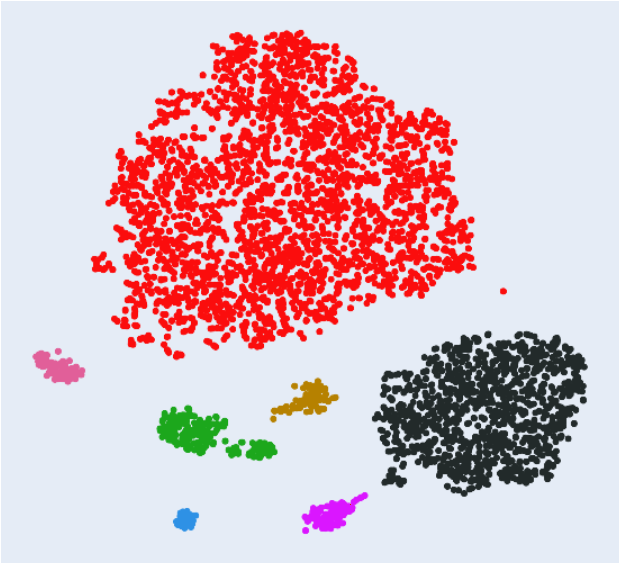}
	\end{subfigure}
	\caption{(left) t-SNE dimensionality reduction on the cytometry data. Type of cell indicated by color. Obtained from the original paper \cite{cytometry}. (right) ExClus' clustering output on the Cytometry data set with hyperparameters $\alpha$ 250, $\beta$ 1.2.\label{fig:cytometry}}
\end{figure}

Figure \ref{fig:cytometry}(left) shows the results directly obtained from the original research paper. This dimensionality reduced version of the data set, calculated with t-SNE, has colored labels for each data point. The markers indicate which type of cell it is. These results are similar to the ones obtained from ExClus (Figure~\ref{fig:cytometry}, right). The two large clusters are immediately distinguishable as similar and the smaller clusters also align almost perfectly. For example, the blue cluster from the ExClus results corresponds to the purple cluster from the original paper and the magenta cluster corresponds to the blue cluster.

 Furthermore, ExClus simplifies the explanations. In the original cytometry paper color maps, showing the expressions of each gene marker, allows for analyzing the dimensionality reduced clustering. On the other hand, ExClus selects the most important attributes and represents them as a distributions compared to the data set's average. For example, to evaluate the blue cluster using the original method a user would have to examine all nine color maps while the ExClus algorithm only presents two attributes.
 
 This case study shows that ExClus can explain the data set to the same extent as the paper, but with a lower descriptional complexity as a user does not need to scan all the different marker figures, which eventually simplifies and speeds up the process of understanding a data set.

\subsection{Hyperparameters}
Both parameters affect the number of clusters and detailedness of explanations, but they do this differently. $\alpha$ is a startup cost for the description length allowing for a more detailed explanation and more clusters as it results in the need for a larger description length before having a substantial impact on the ratio. This parameter's value has values between zero and a thousand. On the other hand, $\beta$ serves as a penalty for the description length and usually has a value between one and two. If the value is larger than one, this applies a penalty that increases super-linearly with an increasing description length, forcing a faster cutoff on explanation and clustering.

\begin{figure}[tp]
    \centering
    \begin{subfigure}{0.48\linewidth}
        \centering
        \includegraphics[height=52mm]{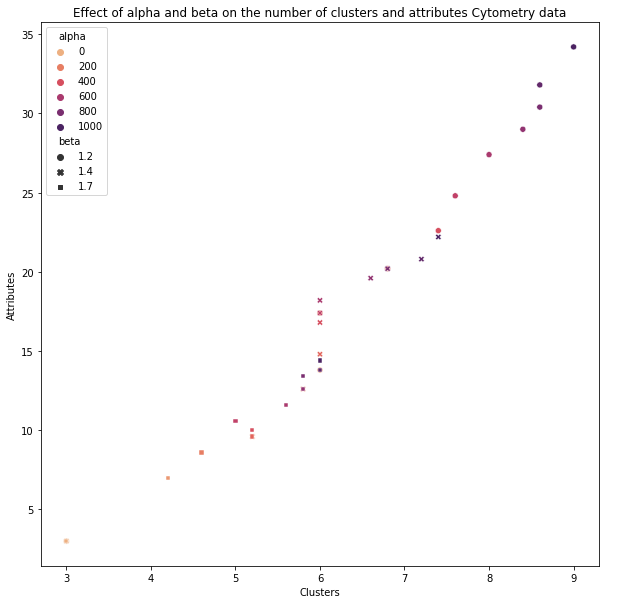}
        \caption{Number of attributes and clusters for various $\alpha$ (color) and $\beta$ (marker symbol) on the Cytometry data.\label{fig:hyper}}
    \end{subfigure}%
    \hfill
    \begin{subfigure}{0.48\linewidth}
        \centering
        \includegraphics[height=52mm]{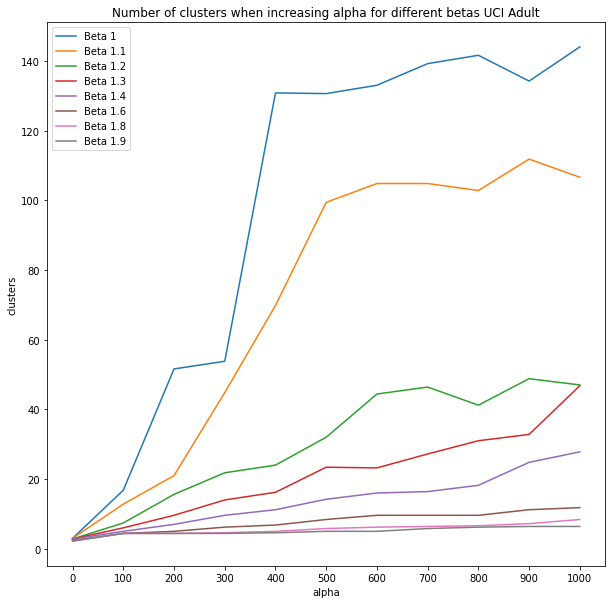}
        \caption{Number of clusters for various $\alpha$ (x-axis) and $\beta$ (colored lines) on the UCI Adult data.\label{fig:hyper_necessary}}
	\end{subfigure}
	\caption{Effects of varying the hyperparameters $\alpha$ and $\beta$.}
\end{figure}

% \begin{figure}[t]
%     \centering
%     \includegraphics[width=0.6\textwidth]{alpha_beta_clusters_attributes_hyperparameters.png}
% 	\caption{Effects of $\alpha$ and $\beta$ on number of clusters (x-axis) and attributes (y-axis) for a Cytometry data set \cite{cytometry}. Difference in $\alpha$ values shown with different colors and different $\beta$ values use different markers.}
% 	\label{fig:hyper}
% \end{figure}

% \begin{figure}[t]
%     \centering
%     \includegraphics[width=.6\textwidth]{n_clusters_alpha_effect_uci.png}
% 	\caption{Number of clusters the ExClus algorithm decided on for different values of $\alpha$ and $\beta$ on the UCI Adultdata set}
% 	\label{fig:hyper_necessary}
% \end{figure}

Figure~\ref{fig:hyper} shows the effect of $\alpha$ and $\beta$ on the number of clusters and attributes. This graph reveals an almost linear increase in the number of clusters and attributes, where $\beta$ divides the values into different sections, and $\alpha$ ensures a further evolution within each section. 
% While this figure only shows results for three different $\beta$'s, which means there will be more overlap in the sections if it includes all values, it does show how the hyperparameters concurrently affect the results.
Furthermore, Figure \ref{fig:hyper_necessary} is a testament to why these parameters are necessary. It presents the number of clusters (y-axis) in the clustering the ExClus algorithm decided on for different values of $\alpha$ (x-axis) and $\beta$ (color). If $\alpha$ is zero, there are almost no clusters, and changing $\beta$ does not affect the results. Besides that, if $\beta$ is one (no penalty), $\alpha$ changes the results almost uncontrollably (blue graph).
While $\alpha$ and $\beta$ might not have a straightforward relationship and interaction, they are both necessary and allow users to change the number of clusters and the number of attributes.

\begin{figure}[t]
    \centering
    \begin{subfigure}{0.48\linewidth}
        \centering
        \includegraphics[width=0.88\textwidth]{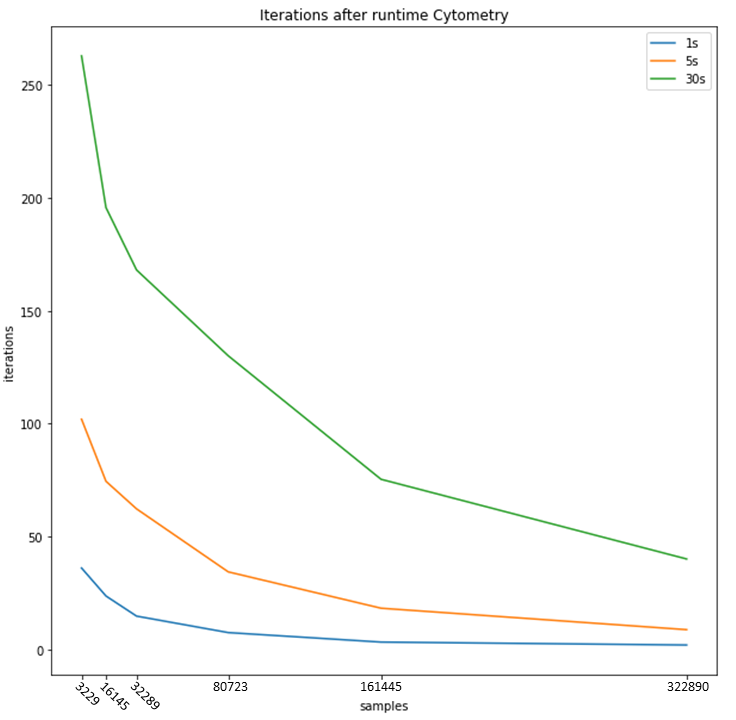}
        \caption{Number of iterations vs. number of \emph{data points} on the Cytometry data.\label{fig:scale_points}}
    \end{subfigure}%
    \hfill
    \begin{subfigure}{0.48\linewidth}
        \centering
        \includegraphics[width=0.88\textwidth]{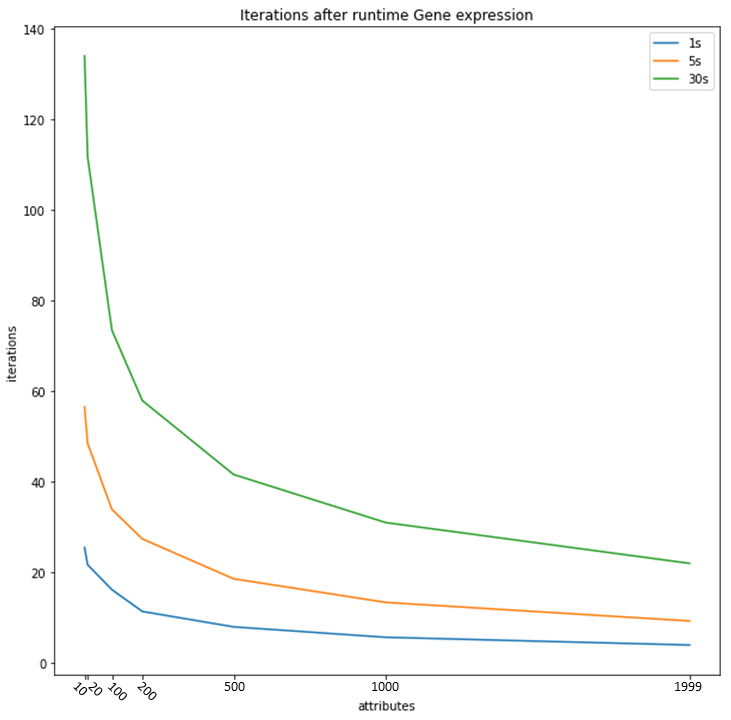}
        \caption{Number of iterations vs. number of \emph{features} on the Gene expression data.\label{fig:scale_attr}}
	\end{subfigure}
	\caption{Number of iterations the algorithm reaches within specified runtimes.}
\end{figure}

\subsection{Scalability}
Data sets can scale in two dimensions. On the one hand, the number of data points can increase, and on the other hand, the number of features can increase. Both can have severe effects on the runtime, and they should be taken into consideration whenever the algorithm runs.
The algorithm goes through multiple iterations of the greedy optimization step. Each iteration searches for the optimal clustering with the number of clusters equal to the iteration. When the runtime ends, it selects the iteration's outcome with the highest subjective interestingness. Therefore, it is best to define scalability in the number of iterations the algorithm reaches. Figure \ref{fig:scale_points} presents these effects when the number of data points increases and Figure \ref{fig:scale_attr} when the number of features increases.

% \begin{figure}[t]
%     \centering
%     \includegraphics[width=.6\textwidth]{iterations_after_runtime_average_cytometry.png}
% 	\caption{Number of iterations the ExClus algorithm reaches after specified runtimes (1s, 5s,30s) with an increasing number of data points on the Cytometry data set.}
% 	\label{fig:scale_points}
% \end{figure}

% \begin{figure}[t]
%     \centering
%     \includegraphics[width=.6\linewidth]{iterations_after_runtime_genex_average.png}
% 	\caption{Number of iterations the ExClus algorithm reaches after specified runtimes (1s, 5s,30s) with an increasing number of features on a Gene expression data set.}
% 	\label{fig:scale_attr}
% \end{figure}

While both figures show an exponential decrease in iterations, this only becomes a noticeable issue if the data set is enormous (more than 100,000 data points or more than 500 features). Furthermore, in most cases, some sampling or feature importance selection will occur if the data set is this large because dimensionality reduction methods and hierarchical clustering techniques can also suffer from scalability issues. Besides that, it is unlikely for these data sets that a clustering with a vast number of clusters will be the optimal result as understanding more than a hundred patterns is highly complex. However, while it is not a limiting issue in most cases, the algorithm does have room for improvement in certain areas that could increase performance.

\section{Conclusion\label{sec:conclusions}}
This paper introduced a new method to assist users in analyzing high-dimensional data sets. It contributes to the state of the art in considering how to find a good balance between informativeness and the complexity of information presented to the user, by making a trade-off between information and its descriptional complexity. Specifically, while generating clusterings and their accompanied optimal explanations automatically. The presented algorithm to identify an explainable clustering on top of a scatter plot of dimensionality-reduced data is implemented in a publicly-available tool called ExClus.

From case studies we have observed that ExClus can be used to effectively analyse data, by comparing results with previous studies on that data (German SE and Cytometry data), highlighting ExClus enables identification of previously known and possibly new patterns with little effort.

Further study could include investigation in methods to choose or support users in choosing good values for the hyperparameters. ExClus allows quick experimentation with hyperparameters, but for now it remains an exercise of trial-and-error for users of the system. Secondly, we have omitted from the scope of this study which dimensionality-reduction method to use. We have used t-SNE in all experiments, which is not the most scalable algorithm and about which many critiques have been written. It would be worthwhile to investigate which dimensionality-reduction methods would synergize best with ExClus. Finally, we have observed that the linkage criterion can be important to consider. More generally, it may be interesting to study other ways to build a restricted search space like the agglomerative clustering approach used here.

\bibliographystyle{splncs04}
\bibliography{references}

\end{document}